\documentclass[review]{elsarticle}

\usepackage{lineno}

\usepackage{times}
\usepackage{helvet}
\usepackage{courier}
\usepackage{amssymb}
\usepackage{latexsym}
\usepackage{amsmath}
\usepackage{multirow}
\usepackage{url}
\usepackage{amsmath}
\usepackage{epsf}
\usepackage{epsfig}
\usepackage{graphicx}
\usepackage{color}
\usepackage{wrapfig}
\usepackage{colortbl}
\usepackage{verbatim}
\usepackage{subfigure}
\usepackage{multirow,rotating}
\usepackage{xcolor} 
\usepackage{pdfpages}

\usepackage{booktabs}
\usepackage{times}  
\usepackage{helvet}  
\usepackage{courier}  
\usepackage{url}  
\usepackage{graphicx}  
\usepackage{amsmath,graphicx,subfigure,color}
\usepackage{algorithm}
\usepackage{algpseudocode}
\usepackage{amssymb}
\usepackage{amsmath}
\usepackage{lscape}
\usepackage{multirow}
\usepackage{hyperref}

\usepackage[normalem]{ulem}
\usepackage[outercaption]{sidecap}    
\usepackage[listings]{tcolorbox}

\modulolinenumbers[5]

\journal{Neurocomputing}

\begin{document}

\begin{frontmatter}

\title{Ensemble-Based Deep Reinforcement Learning for Chatbots}

\author{Heriberto Cuay\'ahuitl$^a$\footnote{Work carried out while the first author was visiting Samsung Research.}, Donghyeon Lee$^b$, Seonghan Ryu$^b$, Yongjin Cho$^b$, Sungja Choi$^b$, Satish Indurthi$^b$, Seunghak Yu$^b$, Hyungtak Choi$^b$, Inchul Hwang$^b$, Jihie Kim$^b$}
\address{$^a$University of Lincoln, School of Computer Science,\\
Lincoln Centre for Autonomous Systems (L-CAS),\\
Brayford Pool, Lincoln, LN6 7TS, United Kingdom}
\address{$^b$Samsung Research, Artificial Intelligence Group,\\
56 Seongchon-gil, Yangjae, Seocho-gu, Seoul, South Korea}




\begin{abstract}
\textcolor{black}{Trainable chatbots that exhibit fluent and human-like conversations remain a big challenge in artificial intelligence. Deep Reinforcement Learning (DRL) is promising for addressing this challenge, but its successful application remains an open question.} This article describes a novel ensemble-based approach applied to value-based DRL chatbots, which use finite action sets as a form of meaning representation. In our approach, while dialogue actions are derived from sentence clustering, the training datasets in our ensemble are derived from dialogue clustering. The latter aim to induce specialised agents that learn to interact in a particular style. In order to facilitate neural chatbot training using our proposed approach, we assume dialogue data in raw text only -- without any manually-labelled data. \textcolor{black}{Experimental results using chitchat data reveal that (1) near human-like dialogue policies can be induced, (2) generalisation to unseen data is a difficult problem, and (3) training an ensemble of chatbot agents is essential for improved performance over using a single agent. In addition to evaluations using held-out data, our results are further supported by a human evaluation that rated dialogues in terms of fluency, engagingness and consistency -- which revealed that our proposed dialogue rewards strongly correlate with human judgements.\footnote{https://doi.org/10.1016/j.neucom.2019.08.007}}
\end{abstract}

\begin{keyword}
Deep Supervised/Unsupervised/Reinforcement Learning, Neural Chatbots
\end{keyword}

\end{frontmatter}


\section{Introduction}
\label{intro}
Humans in general find it relatively easy to have chat-like conversations that are both coherent and engaging at the same time. While not all human chat is engaging, it is arguably coherent \cite{GroszS86}, and it can cover  large vocabularies across a wide range of conversational topics. In addition, each contribution by a partner conversant may exhibit multiple sentences, such as greeting+question or acknowledgement+statement+question. The topics raised in a conversation may go back and forth without losing coherence. All of these phenomena represent big challenges for current data-driven chatbots.

\begin{figure*}[t!]
\begin{center}
\includegraphics[width=12.2cm]{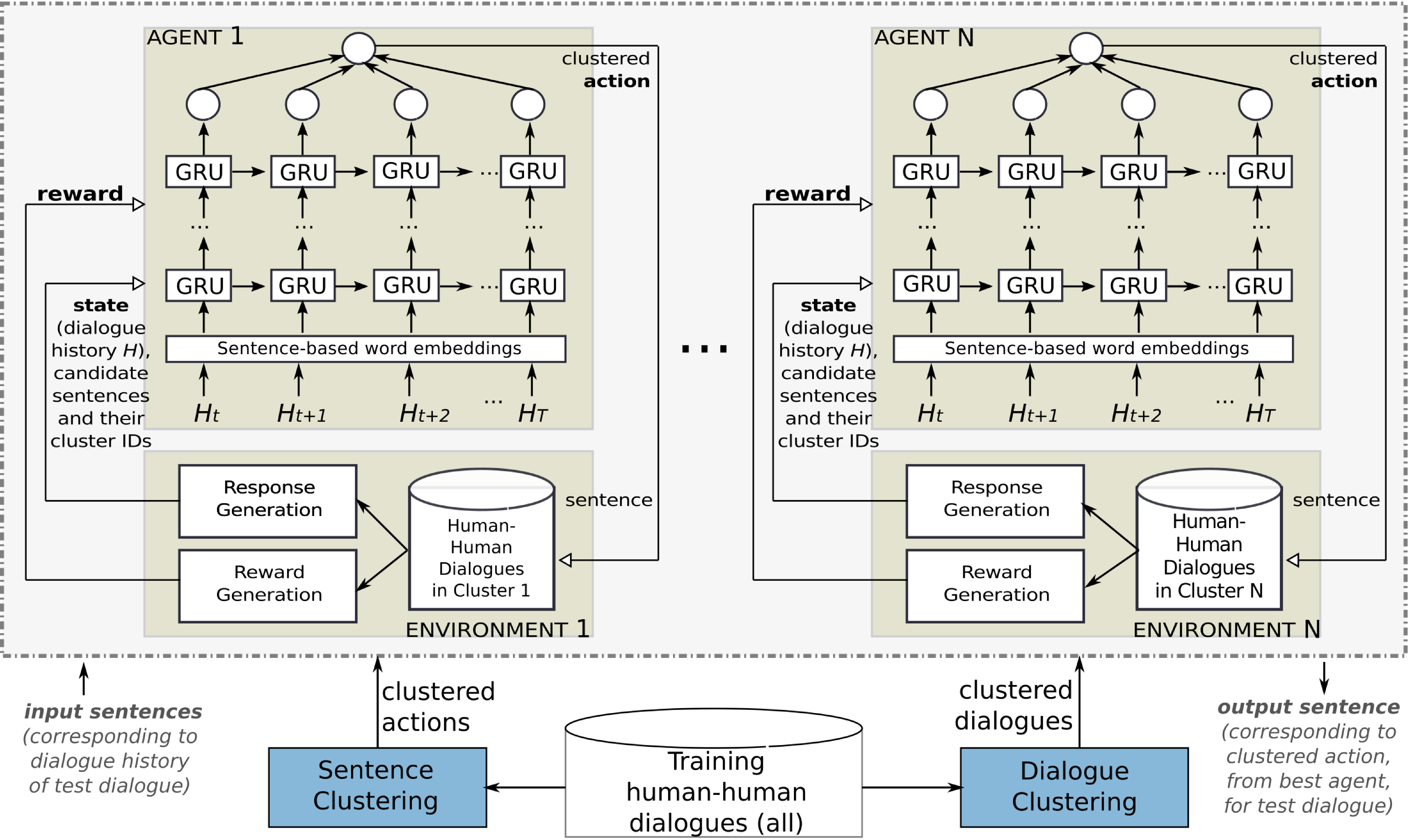}
\end{center}
\caption{High-level architecture of the proposed ensemble-based learning approach for chatbot training---see text for details}
\label{agent-arch}
\end{figure*}

\textcolor{black}{We present a novel approach for chatbot training based on the reinforcement learning \cite{SuttonB2018}, unsupervised learning \cite{HastieTF09} and deep learning \cite{LeCunBH15} paradigms. In contrast to other learning approaches for Deep Reinforcement Learning chatbots that rely on partially labelled dialogue data \cite{SerbanEtAl2018,LiMSJRJ16}, our approach assumes only unlabelled data.} Our learning scenario is as follows: given a dataset of human-human dialogues in raw text (without any manually provided labels), an ensemble of Deep Reinforcement Learning (DRL) agents take the role of one of the two partner conversants in order to learn to select human-like sentences when exposed to both human-like and non-human-like sentences. In our learning scenario the agent-environment interactions consist of agent-data interactions -- there is no user simulator as in task-oriented dialogue systems \cite{Cuayahuitl16,GaoGL19}. During each verbal contribution and during training, the DRL agents 
\begin{enumerate}
\itemsep0em 
\item observe the state of the world via a recurrent neural network, which models a representation of all words raised in the conversation together with a set of candidate responses (i.e. {\it clustered actions} in our approach); 
\item they then select an action so that their word-based representation is sent to the environment; and 
\item they receive an updated dialogue history and a numerical reward for having chosen a certain action, until a termination condition is met. 
\end{enumerate}
This process---illustrated in Figure~\ref{agent-arch}---is carried out iteratively until the end of a dialogue for as many dialogues as necessary, i.e. until there is no further improvement in the agents' performance. 
During each verbal contribution at test time, the agent exhibiting the highest predictive dialogue reward is selected for human-agent interactions.

This article makes the following contributions to neural-based chatbots:
\begin{enumerate}
\item We propose a novel approach for chatbot training using  value-based Deep Reinforcement Learning, where we induce action sets automatically via unsupervised clustering. Most previous related work has used policy search methods, and value-based methods have received little attention. We identified the latter as a research gap in our literature review. Although the performance of our DRL agents drops with dialogues that the agents are not familiar with, our DRL agents indeed learn to improve their performance over time with dialogues that they get familiarised with.  
\item We propose a novel reward function due to the lack of well-embraced metrics for measuring chatbot performance. In addition, we train neural regressors for predicting dialogue rewards using a dataset of human-human dialogues that was automatically extended with noisy dialogues. While non-noisy dialogues exemplify human-like and desirable outputs, the noisy ones exemplify less desirable behaviour. This reward function is easy to implement, it strongly correlates with test human-human dialogues subject to using long dialogue histories, and it strongly correlates with human judgements.
\item We propose a novel ensemble-based methodology for chatbot training, where each chatbot in our ensemble is trained with a set of clustered dialogues. To test our agents, we train 100 DRL chatbots with the aim of generating more context-relevant responses. Our experimental results according to automatic and human evaluations show that the ensemble of DRL agents outperforms a single DRL agent. This result is relevant for training future neural-based chatbots.
\end{enumerate}

In the next two sections, 2 and 3, we review related work on neural-based chatbots and provide related background on deep reinforcement learning. Then we describe our proposed approach and methodology in section 4. This is followed by a comprehensive set of automatic and human evaluations in section 5, which use (i) a dataset of chitchat conversations, and (ii) human ratings of human-chatbot dialogues. Section 6 draws conclusions and discusses avenues for future research.

\section{Related Work}
\label{litreview}

\paragraph{Deep Reinforcement Learning Chatbots} Reinforcement Learning (RL) methods are typically based on value functions or policy search \cite{SuttonB2018}, which also applies to deep RL methods. Both types of trained agents can use the same state representations and rewards, but they differ in the representation of actions and policies. While value-based methods are typically applied to problems with discrete and finite actions, policy search methods can be applied to problems with either finite or infinite actions. In addition, while policies in value-based methods calculate numerical values (also referred to as `expected long-term rewards') to model the importance of each state-action pair, policy search methods induce policies directly \cite{Li17b}---via the parameters of a model such as a neural network or a Gaussian process. While value-based methods have been particularly applied to task-oriented dialogue systems \cite{CasanuevaBSURTG18,CuayahuitlY17,CuayahuitlYWC17,WilliamsAZ17,PengLLGCLW17, HC2016iwsds}, policy-based methods have been particularly applied to open-ended dialogue systems such as (chitchat) chatbots \cite{LiMSJRJ16,LiMSJRJ17,SerbanEtAl2018}. This is not surprising given the fact that task-oriented dialogue systems use finite action sets, while chatbot systems use infinite action sets. The latter consider each sentence as an action, and consequently, the task is to induce dialogue behaviour from an infinite action set. This is extremely challenging for value-based reinforcement learning methods, which are more suitable for solving problems with finite action spaces. So far there is a preference for policy search methods for chatbots, but it is not clear whether they should be preferred because they face problems such as local optima rather than global optima, inefficiency, and high variance. It is therefore that we explore the feasibility of value-based methods for chatbots with large action sets, which has not been explored before---especially not from the perspective of deriving the action sets automatically as attempted in this article. 

\paragraph{Sequence2Sequence Chatbots} Other closely related methods to DRL include sequence to sequence models for dialogue generation \cite{VinyalsL15,SordoniGABJMNGD15,SerbanKTTZBC17,LiGBSGD16,Wang2018,ZhangEtAl2018}. \textcolor{black}{These methods are based on Recurrent Neural Networks (RNNs) using an encoder-decoder architecture. In these methods, while one RNN referred to as `encoder' computes an internal representation of the inputs, another RNN referred to as `decoder' generates one word as a time -- both trained end-to-end with all parameters (weights) trained jointly. Given a sequence of input words $(x_1,\cdots,x_T)$, an encoder-decoder computes a sequence of output words $(y_1,\cdots,y_T)$  by iterating the following equation: $y_t=softmax({\bf W}^{yh} {\bf h}_t)$, where $softmax(z_i)=\frac{\exp^z_i}{\sum_j \exp^z_j}$, ${\bf h}_t = \sigma ({\bf W}^{xh} {\bf x}_t+{\bf W}^{hh} {\bf h}_{t-1})$, function $\sigma(.)$ is an activation function, and training weights ${\bf W}$ involves minimising a loss function such as categorical cross entropy \cite{SutskeverVL14}. Sequence2Sequence (Seq2Seq) methods can be combined with deep reinforcement learners by treating the policy as an encoder-decoder \cite{LiMSJRJ16}. Seq2Seq methods have also been included in ensemble-based methods together with rule-based systems and a variety of other machine learning methods \cite{SerbanEtAl2018,SongLNZZY18,PapaioannouEtAl2017}.} While some of them use a single DRL agent \cite{SerbanEtAl2018}, they have not been investigated using an ensemble containing a horde of DRL-based chatbots as attempted in this article. 
 
\paragraph{Reward Functions} Related work above highlights that evaluation is a difficult part, and that there is a need for better evaluation metrics. \textcolor{black}{This is further supported by \cite{LiuLSNCP16}, who found that typical metrics used to assess the quality of machine translators such as BLEU (Bilingual Evaluation Understudy) \cite{PapineniEtAl2002} and METEOR (Metric for Evaluation of Translation with Explicit ORdering) \cite{Lavie2007} amongst others do not correlate with human judgments.} The dialogue rewards used by DRL agents are either specified manually depending on the application, or learnt from dialogue data. For example, \cite{LiMSJRJ16} conceives a reward function that positively rewards sentences that are easy to respond to and coherent while penalising repetitiveness. \cite{LiMSJRJ17} uses an adversarial approach, where the discriminator is trained to score human vs. non-human sentences so that the generator can use these scores during training. \cite{SerbanEtAl2018} trains a reward function from expensive and time-consuming human ratings. All these related studies are neural-based, and there is no clear best reward function to use in future (chitchat) chatbots. This motivated us to propose a new metric that is easy to implement, practical due to requiring only data in raw text, and potentially promising as described below.

\textcolor{black}{This article contributes to the literature of neural-based chatbots as follows. First, our methodology for training value-based DRL agents uses only unlabelled dialogue data. Previous work requires manual extensions to the dialogue data \cite{LiMSJRJ16} or expensive and time consuming ratings for training a reward function \cite{SerbanEtAl2018}. Second, our proposed reward function strongly correlates with human judgements. Previous work has only shown moderate positive correlations between target dialogue rewards and predicted ones \cite{SerbanEtAl2018}, or rely on high-level annotations requiring external and language-dependent resources typically induced from labelled data \cite{DHaroBHL19}. Third, while previous work on DRL chatbots train a single agent \cite{SerbanEtAl2018,LiMSJRJ16}, our study---confirmed by automatic and human evaluations---shows that an ensemble-based approach performs better than a counterpart single agent. The remainder of this article elaborates on these contributions.}

\section{Background}
\label{background}
A reinforcement learning agent induces its behaviour from interacting with an environment through trial and error, where situations (representations of sentences in a dialogue history) are mapped to actions (follow-up sentences) by maximising a long-term reward signal. Such an agent is typically characterised by: 
(i) a finite set of states $S=\{s_i\}$ that describe all possible situations in the environment; 
(ii) a finite set of actions $A=\{a_j\}$ to change in the environment from one situation to another; 
(iii) a state transition function $T(s,a,s')$ that specifies the next state $s'$ for having taken action $a$ in the current state $s$; 
(iv) a reward function $R(s,a,s')$ that specifies a numerical value given to the agent for taking action $a$ in state $s$ and transitioning to state $s'$; and 
(v) a policy $\pi:S \rightarrow A$ that defines a mapping from states to actions \cite{SuttonB2018,Szepesvari:2010}. 

The goal of a reinforcement learning agent is to find an optimal policy by maximising its cumulative discounted reward defined as 
\begin{equation}\nonumber
Q^*(s,a)=\max_\pi \mathbb{E}[r_t+\gamma r_{t+1}+\gamma^2 r_{t+1}+...|s_t=s,a_t=a,\pi],
\end{equation}
where function $Q^*$ represents the maximum sum of rewards $r_t$ discounted by factor $\gamma$ at each time step. 
While a reinforcement learning agent takes actions with probability $Pr(a|s)$ during training, it selects the best action at test time according to 
\begin{equation}\nonumber
\pi^{*}(s)=\arg \max_{a \in A} Q^*(s,a).
\end{equation}

A deep reinforcement learning agent approximates $Q^*$ using a multi-layer neural network \cite{MnihKSRVBGRFOPB15}. The $Q$ function is parameterised as $Q(s,a;\theta)$, where $\theta$ are the parameters or weights of the neural network (recurrent neural network in our case). Estimating these weights requires a dataset of learning experiences $D=\{e_1,...e_N\}$ (also referred to as `experience replay memory'), where every experience is described as a tuple  $e_t=(s_t,a_t,r_t,s_{t+1})$. Inducing a $Q$ function consists in applying Q-learning updates over minibatches of experience $MB=\{(s,a,r,s')\sim U(D)\}$ drawn uniformly at random from the full dataset $D$. 
\textcolor{black}{This process is implemented in learning algorithms using Deep Q-Networks (DQN) such as those described in \cite{MnihKSRVBGRFOPB15,HasseltGS16,WangSHHLF16}, and the following section describes a DQN-based algorithm for human-chatbot interaction.}

\section{Proposed Approach}
\label{approach}

This section explains the main components of Figure~\ref{agent-arch} as follows. Motivated by \cite{CuayahuitlEtAl2019ijcnn}, we first describe the ensemble of Deep Reinforcement Learning (DRL) agents, we then explain how to conceive a finite set of dialogue actions from raw text, and finally we describe how to assign dialogue rewards for training DRL-based chatbots. 

\subsection{Ensemble of DRL Chatbots}
\label{ensemble}
We assume that all deep reinforcement learning agents in our ensemble use the same neural network architecture and learning algorithm. They only differ in the portion of data used for training and consequently the weights in their trained models---see \cite{Wiering2008EAR,ChenEtAl2018} for alternative approaches. 
Our agents aim to maximise their cumulative reward over time according to 
\begin{equation}\nonumber
Q^*(s,a;\theta_i)=\max_{\pi_{\theta_i}} {\mathbb E}[r_t+\gamma r_{t+1}+\gamma^2 r_{t+2}+\cdots|s,a,\pi_{\theta_i}], 
\end{equation}
where $r$ is the numerical reward given at time step $t$ for choosing action $a$ in state $s$, $\gamma$ is a discounting factor, and $Q^*(s,a;\theta_i)$ is the optimal action-value function using weights $\theta$ in the neural network of chatbot $i$. During training, a DRL agent will choose actions in a probabilistic manner in order to explore new $(s,a)$ pairs for discovering better rewards or to exploit already learnt values---with a reduced level of exploration over time and an increased level of exploitation over time. During testing, our ensemble-based DRL chatbot will choose the best actions $a^*$ according to 
\begin{equation}\nonumber
\pi^*(s)=\arg\max_{a \in A} Q^*(s,a;\theta_i) \mbox{ with } i=\arg\max_{i \in I} \hat{R}(\tau_i),
\end{equation}
where $\tau_i=<(s_0,a_0),...,(s_t,a_t)>$ is a trajectory of state-action pairs of chatbot $i$, and $\hat{R}(\tau_i)$ is a function that predicts the dialogue reward of chatbot $i$ as in \cite{CuayahuitlEtAl2018neurips}. Given the set of trajectories for all agents---where each agent takes its own decisions and updates its environment states accordingly---the agent with the highest predictive reward is selected, i.e. the one with the least amount of errors in the interaction. 

Our DRL agents implement the procedure above using a generalisation of DQN-based methods \cite{MnihKSRVBGRFOPB15,HasseltGS16,WangSHHLF16}---see Algorithm~\ref{ChatDQN}, explained as follows. 
\begin{itemize}
\item After initialising replay memory $D=\{e_1,\dots,e_{|D|}\}$ with learning experience $e_i=(s,a,r,s')$, dialogue history $H=\{s_1,\dots,s_{|H|}\}$ with sentences $s_i$, action-value function $Q$ and target action-value function $\hat{Q}$, we sample a training dialogue from our data of human-human conversations (lines 1-4). 
\item Once a conversation starts, it is mapped to its corresponding sentence embedding representation, i.e. `sentence vectors' as described in Section~\ref{clusteredactions} (lines 5-6). 
\item Then a set of candidate responses is generated including (1) the true human response and (2) a set of randomly chosen responses (distractors). The candidate responses are clustered as described in the next section and the resulting actions are taken into account by the agent for action selection (lines 8-10). 
\item Once an action is chosen, it is conveyed to the environment, a reward is observed as described at the end of this section, and the agent's partner response is observed in order to update the dialogue history $H$ (lines 11-14). 
\item In response to the update above, the new sentence embedding representation is extracted from $H$ for updating the replay memory $D$ with experience $e$ (lines 15-16). 
\item Then a minibatch of experiences $MB=\{e_j\}$ is sampled from $D$ for updating weights $\theta$ according to the error derived from the difference between the target value $y_j$ and the predicted value $Q(s,a;\theta)$ (see lines 18 and 20), which is based on the following weight updates:
\begin{equation}\nonumber
\theta_{t'}=\theta_{t} + \alpha (y_t - Q(s,a;\theta_t)) \nabla_{\theta_t} Q(s,a;\theta_t),
\end{equation}
where $y_t=r_{t} + \gamma \max_{a'} \hat{Q}(s',a';\hat{\theta_t})$ and $\alpha$ is a learning rate hyperparameter. 
\item The target action-value function $\hat{Q}$ and environment state $s$ are updated accordingly (lines 21-22), and this iterative procedure continues until convergence.
\end{itemize}

\begin{algorithm} 
\caption{\label{ChatDQN} ChatDQN Learning}\label{ndqn} 
\begin{algorithmic}[1]
\State Initialise Deep Q-Networks with replay memory $D$, dialogue history $H$,  action-value function $Q$ with random weights $\theta$, and target action-value functions $\hat{Q}$ with $\hat{\theta}=\theta$
\State Initialise clustering model from training dialogue data
\Repeat
   \State Sample a training dialogue (human-human sentences)
   \State Append first sentence to dialogue history $H$
   \State $s=$ sentence embedding representation of $H$
   \Repeat 
      \State Generate {\it noisy} candidate response sentences
      \State $A=\text{cluster IDs of candidate response sentences}$
      \State $a= 
\begin{cases}
    rand_{a \in A} \text{ if } \mbox{random number} \le \epsilon\\
    \max_{a \in A} Q(s,a;\theta)  \text{  otherwise}
\end{cases}$
      \State Execute chosen clustered action $a$
      \State Observe human-likeness dialogue reward $r$
      \State Observe environment response (agent's partner)
      \State Append agent and environment responses to $H$
      \State $s'=$ sentence embedding representation of $H$
      \State Append learning experience $e=(s,a,r,s')$ to $D$
      \State Sample random minibatch $(s_j,a_j,r_j,s'_j)$ from $D$
      \State $y_j= 
\begin{cases}
    r_j \text{ if } \mbox{final step of episode}\\
    r_j + \gamma \max_{a' \in A} \hat{Q}(s',a';\hat{\theta})              & \text{otherwise}
\end{cases}$
      \State Set $err=\left( y_j-Q(s,a;\theta) \right)^2$ 
      \State Gradient descent step on $err$ with respect to $\theta$
      \State Reset $\hat{Q}=Q$ every $C$ steps
      \State $s \leftarrow$ $s'$
   \Until {end of dialogue}
   \State Reset dialogue history $H$
\Until convergence
\end{algorithmic}
\end{algorithm}

\subsection{Sentence and Dialogue Clustering}
\label{clusteredactions}
Actions in reinforcement learning chatbots correspond to sentences, and their size is infinite assuming all  possible combinations of word sequences in a given language. This is especially true in the case of open-ended conversations that make use of large vocabularies, as opposed to task-oriented conversations that make use of smaller (restricted) vocabularies. A {\bf clustered action} is a group of sentences sharing a similar or related meaning via {\it sentence vectors} derived from word embeddings \cite{MikolovSCCD13,PenningtonSM14}. We represent sentences via their mean word vectors---similarly as in Deep Averaging Networks \cite{IyyerMBD15}---denoted as ${\bf x}_l=\frac{1}{N_l}\sum_{j=1}^{N_l} c_j$, where $c_j$ is the vector of coefficients of word $j$, $N_l$ is the number of words in sentence $l$, and ${\bf x}_l$ is the embedding vector of sentence $l$. Similarly, a {\bf clustered dialogue} is a group of conversations sharing a similar or related topic(s) via their clustered actions. We represent dialogues via their clustered actions. Dialogue clustering in this way can be seen as a two-stage approach, where sentences are clustered in the first step and dialogues are clustered in the second step. In our proposed approach, each DRL agent is trained on a cluster of dialogues.

While there are multiple ways of selecting features for clustering and also multiple clustering algorithms, the following requirements arise for chatbots: (1) unlabelled data due to human-human dialogues in raw text (this makes it difficult to evaluate the goodness of clustering features and algorithms), and (2) scalability to clustering a large set of data points (especially in the case of sentences, which are substantially different between them due to their open-ended nature).

Given a set of data points $\{{\bf x}_1,\cdots,{\bf x}_n\} \forall {\bf x}_l \in \mathbb{R}^m$ and a similarity metric 
$d({\bf x}_l,{\bf x}_{l'})$, the task is to find a set of $\mathcal{K}$ groups with a clustering algorithm. In our case each data point ${\bf x}$ corresponds to a dialogue or a sentence. For scalability purposes, we use the K-Means++ algorithm \cite{ArthurV07} and the Euclidean distance $d({\bf x}_l,{\bf x}_{l'})=\sqrt{\sum_{j=1}^m({x}_l^j-{x}_{l'}^j)^2}$ with $m$ dimensions, and consider $\mathcal{K}$ as a hyperparameter -- though other clustering algorithms and distance metrics can be used with our approach. In this way, a trained sentence clustering model assigns a cluster ID $a \in A$ to features ${\bf x}_l$, where the number of actions (in a DRL agent) refers to the number of sentence clusters, i.e. $\vert A \vert = \mathcal{K}$.

\subsection{Human-Likeness Rewards}
\label{humanrewards}
Specifying reward functions in reinforcement learning dialogue agents is often a difficult aspect. We propose to derive rewards from human-human dialogues by assigning positive values to contextualised responses seen in the data, and negative values to randomly chosen responses due to lacking coherence (also referred to as `non-human-like responses') -- see example in Tables~\ref{exampledialogue_good} and \ref{exampledialogue_bad}. An episode or dialogue reward can thus be computed as $R_d=\sum_{j=1}^N r_j^d(a)$, where index $d$ refers to the dialogue in focus, index $j$  to the dialogue turn in focus, and $r_j^d(a)$ is given according to
\begin{equation}\nonumber
  r^d_j(a)=\begin{cases}
    +1, & \text{if $a$ is a human response in dialogue-turn $j$ of dialogue $d$.}\\
    -1, & \text{if $a$ is human but randomly chosen (incoherent)}.
  \end{cases}
\end{equation}
Table~\ref{exampledialogue_good} shows an example of a well rewarded dialogue (without distortions) and Table~\ref{exampledialogue_bad} shows an example of a poorly rewarded dialogue (with distortions). Other dialogues can exhibit similar dialogue rewards or something in between (ranging between $-T$ and $T$), depending on the amount of distortions---the higher the amount of distortions the lower the dialogue reward. 

\begin{table}[t!]
\fontsize{8.5}{7.5} \selectfont 
  \begin{center}
    \begin{tabular}{|c|l|c|}
\hline
turn & Verbalisation & Reward\\
\hline
\multirow{2}{*}{1} & A: hello what are doing today ? &\\
 & B: i'm good , i just got off work and tired , i have two jobs . & +1\\
\hline
\multirow{2}{*}{2} & A: i just got done watching a horror movie&\\
 & B: i rather read , i have read about 20 books this year . & +1\\
\hline
\multirow{2}{*}{3} & A: wow ! i do love a good horror movie . loving this cooler weather&\\
 & B: but a good movie is always good .& +1\\
\hline
\multirow{2}{*}{4} & A: yes ! my son is in junior high and i just started letting him watch them too&\\
 & B: i work in the movies as well .& +1\\
\hline
\multirow{2}{*}{5} & A: neat ! ! i used to work in the human services field&\\
 & B: yes it is neat , i stunt double , it is so much fun and hard work . & +1\\
\hline
\multirow{2}{*}{6} & A: yes i bet you can get hurt . my wife works and i stay at home&\\
 & B: nice , i only have one parent so now i help out my mom . & +1\\
\hline
\multirow{2}{*}{7} & A: i bet she appreciates that very much .&\\
 & B: she raised me right , i'm just like her . & +1\\
\hline
\multirow{2}{*}{8} & A: my dad was always busy working at home depot&\\
 & B: now that i am older home depot is my toy r us .& +1\\
\hline
    \end{tabular}
  \caption{Example human-human dialogue with a good numerical reward $\sum_{j=1}^{T=8} r_j^1(a)=8$}
  \label{exampledialogue_good}
  \end{center}
\end{table}

\begin{table}[h!]
\fontsize{8.5}{7.5} \selectfont 
  \begin{center}
    \begin{tabular}{|c|l|c|}
\hline
Turn & Verbalisation & Reward\\
\hline
\multirow{3}{*}{1} & A: hello what are doing today ? &\\
 & B: \sout{i'm good , i just got off work and tired , i have two jobs .} & \sout{+1}\\
 & B: lol. she is more pop now. her old stuff was country. i like classic country . & -1\\
\hline
\multirow{3}{*}{2} & A: i just got done watching a horror movie&\\
 & B: \sout{i rather read , i have read about 20 books this year .} & \sout{+1}\\
 & B: i am good how about you & -1\\
\hline
\multirow{3}{*}{3} & A: wow ! i do love a good horror movie . loving this cooler weather&\\
 & B: \sout{but a good movie is always good .}& \sout{+1}\\
 & B: i do not enjoy the driving in my honda civic & -1\\
\hline
\multirow{3}{*}{4} & A: yes ! my son is in junior high and i just started letting him watch them too&\\
 & B: \sout{i work in the movies as well .}& \sout{+1}\\
 & B: do you like dogs ? I have two .& -1\\
\hline
\multirow{3}{*}{5} & A: neat ! ! i used to work in the human services field&\\
 & B: \sout{yes it is neat , i stunt double , it is so much fun and hard work .} & \sout{+1}\\
 & B: very cool . i had one of those . & -1\\
\hline
\multirow{3}{*}{6} & A: yes i bet you can get hurt . my wife works and i stay at home&\\
 & B: \sout{nice , i only have one parent so now i help out my mom .} & \sout{+1}\\
 & B: that is just great . we need more educators like you in the system . & -1\\
\hline
\multirow{3}{*}{7} & A: i bet she appreciates that very much .&\\
 & B: \sout{she raised me right , i'm just like her .} & \sout{+1}\\
 & B: is it bad that i hate pumpkins and trees ? & -1\\
\hline
\multirow{3}{*}{8} & A: my dad was always busy working at home depot&\\
 & B: \sout{now that i am older home depot is my toy r us .}& \sout{+1}\\
 & B: why not ? maybe you do not like to travel ? & -1\\
\hline
    \end{tabular}
  \caption{Example distorted human-human dialogue with a poor numerical reward $\sum_{j=1}^8 r_j^2(a)=-8$}
  \label{exampledialogue_bad}
  \end{center}
\end{table}

We employ the algorithm described in \cite{CuayahuitlEtAl2018neurips} for generating dialogues with varying amounts of distortions (i.e. different degrees of human-likeness), which we use for training and testing reward prediction models using supervised regression. Given our extended dataset $\mathcal{\hat{D}}=\{(\hat{d}_1,y_1),\dots,(\hat{d}_N,y_N)\}$ with (noisy) dialogue histories $\hat{d}_i$ represented with sequences of sentence vectors, the goal is to predict dialogue scores $y_i$ as accurately as possible.

Alternative and automatically derived values between {-1} and {+1} are also possible but considered as future work. Section \ref{sec:human-eval} provides an evaluation of our reward function and its correlation with human judgement. We show that albeit simple, our reward function is highly correlated with our judges' ratings.

\subsection{Methodology}
Our proposed approach can be summarised through the following methodology:
\begin{tcolorbox}
\begin{enumerate}
\item \textcolor{black}{Collect or adopt a dataset of human-human dialogues (as in \ref{data})
\vskip10pt
\item Design or adopt a suitable reward function (as in \ref{humanrewards})
\vskip10pt
\item Train a neural regressor for predicting dialogue rewards (as in \cite{CuayahuitlEtAl2018neurips})
\vskip10pt
\item Perform sentence and dialogue clustering in order to define the action set and training datasets (as in \ref{clusteredactions})
\vskip10pt
\item Train a Deep Reinforcement Learning agent per dialogue cluster (as described in \ref{ensemble})
\vskip10pt
\item Test the ensemble of agents together with the predictor of dialogue rewards (as in \ref{sec:automatic-eval} and \ref{sec:human-eval}), and iterate from Step 1 if needed
\vskip10pt
\item Deploy your trained chatbot subject to satisfactory results in Step 6}
\end{enumerate}
\end{tcolorbox}

\section{Experiments and Results}
\label{experiments}

\subsection{Data}
\label{data}
We used the {\it Persona-Chat} dataset\footnote{Dataset downloaded from \url{http://parl.ai/} on 18 May 2018 \cite{MillerFBBFLPW17}}, stats are shown in Table~\ref{datastats}.

\begin{table}[b!]
\label{datastats}
\footnotesize
\centering
\caption{Statistics of the Persona-Chat data used in our experiments for chatbot training}
\begin{tabular}{|l|c|c|} 
\hline
Attribute / Value & Training Set & Test Set \\
\hline
\hline
Number of dialogues & 17877 & 999\\
Number of dialogue turns & 131438 & 7801\\
Number of sentences & 262862 & 15586\\
Number of unique sentences & 124469 & 15186\\
Number of unique words & 18672 & 6692\\
Avg. turns per dialogue	& 7.35 & 7.8\\
Avg. words per dialogue	& 165.89 & 185.86\\
Avg. words per sentence	& 11.28 & 11.91\\
\hline 
\end{tabular}
\end{table}




\subsection{Experimental Setting} 
Our agents' states model dialogue histories as sequences of sentence vectors---using GloVe-based \cite{PenningtonSM14} mean word vectors \cite{IyyerMBD15}---with pre-trained embeddings. 
All our experiments use a 2-layer Gated Recurrent Unit (GRU) neural network\footnote{Other hyperparameters include embedding batch size=128, dropout=0.2, sentence vector dimension=100, latent dimensionality=256, discount factor=0.99, size of candidate responses=20, max. number of mean sentence vectors in $H$=25, burning steps=3K, memory size=10K, target model update (C)=10K, optimiser=Adam, learning steps=\{100K, 500K\}, test steps=100K. The number of parameters of each neural net is 2.53 million.}  \cite{choEtAlEMNLP2014}. 
At each time step $t$ in the dialogue history, the first hidden layer generates a hidden state ${\bf h}_t$ as follows:
\begin{equation}\nonumber
\begin{gathered}
{\bf r}_t=\sigma ({\bf W}_r {\bf x}_t + {\bf U}_r {\bf h}_{t-1}),\\
{\bf z}_t=\sigma ({\bf W}_z {\bf x}_t + {\bf U}_z {\bf h}_{t-1}),\\
\bar{\bf h}_t=\mbox{tanh} ({\bf W}_{\bar{\bf h}} {\bf x}_t + {\bf U}_{\bar{\bf h}} ({\bf r}_{t} \odot {\bf h}_{t-1})),\\
{\bf h}_{t}=(1-{\bf z}_{t}) \odot {\bf h}_{t-1} + {\bf z}_{t} \odot \bar{{\bf h}_{t}},
\end{gathered}
\end{equation}
where ${\bf x}_t$ refers to a set of sentence vectors of the dialogue history, ${\bf r}_t$ is a reset gate that decides how much of the previous state to forget, ${\bf z}_t$ is an update gate that decides how much to update its activation, $\bar{\bf h}_t$ is an internal state, $\sigma(.)$ and $tahn(.)$ are the Sigmoid and hyperbolic Tangent functions (respectively), ${\bf W}_{*}$ and ${\bf U}_{*}$ are learnt weights, and $\odot$ refers to the element-wise multiplication. If the equations above are summarised as ${\bf h}_{t}=GRU({\bf x}_t,{\bf h}_{t-1})$ we get the following output action taking into account both hidden layers in our neural net: $a_{t}=\arg \max_{a \in A}({\bf W}_{a} {\bf h}_{t}^2)$, where ${\bf h}_{t}^2=GRU({\bf h}_{t}^1,{\bf h}_{t-1}^2)$ and ${\bf h}_{t}^1=GRU({\bf x}_t,{\bf h}_{t-1}^1)$.

\begin{table}[t!]
{\fontsize{10}{9} \selectfont 
\centering
\caption{Example clustered sentences chosen arbitrarily}
\begin{tabular}{|c|l|} 
\hline
ID & Clustered Sentence \\
\hline
\hline
 & `i mostly eat a fresh and raw diet , so i save on groceries .', `i  \\
 & only eat kosher foods', `i like kosher salt a lot', `i prefer seafood . \\
25 & my dad makes awesome fish tacos .', `i do a pet fish', 'that sounds \\ 
 & interesting i like organic foods .', `cheeseburgers are great , i try\\
 & all kinds of foods everywhere i go , gotta love food .'\\
\hline
 & `hi how are you doing ? i am okay how about you ?', `i am great . \\
 & what do you like to do ?', `oh right how i am between jobs', \\
68 & `i am thinking about my upcoming retirement . how about you ?', \\
 & `i am not sure ? how old are you  ?', `i am well , how are you ?'\\
 & `i am doing very fine this afternoon . how about you ?'\\
\hline
 & `i have dogs and i walk them . and a cat .', `yeah dogs are \\
 & pretty cool', `i have dogs and cats', `hello , leon . my dogs and \\
88 & i are doing well .', `sadly , no . my dogs and i are in ohio .', \\
 & `i have 2 dogs . i should take them walking instead of eating .', \\
 & `yeah dogs are cool . i kayak too . do you have pets ?'\\
\hline 
\end{tabular}
\label{clusteredsentences}
}
\end{table}


\begin{figure}[h!]
\centering
\subfigure[100 clusters of training sentences]{\includegraphics[width=120mm]{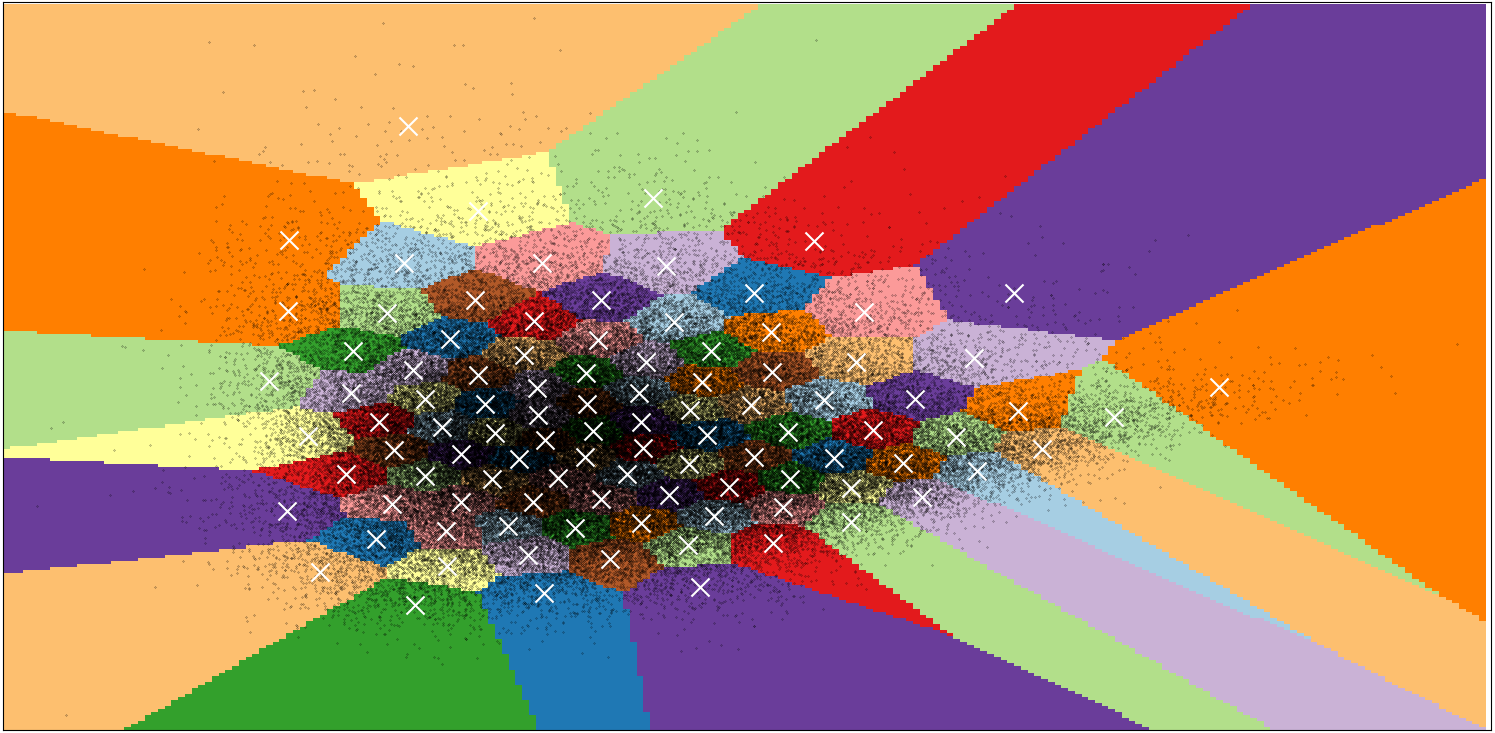}}
\subfigure[100 clusters of training dialogues]{\includegraphics[width=120mm]{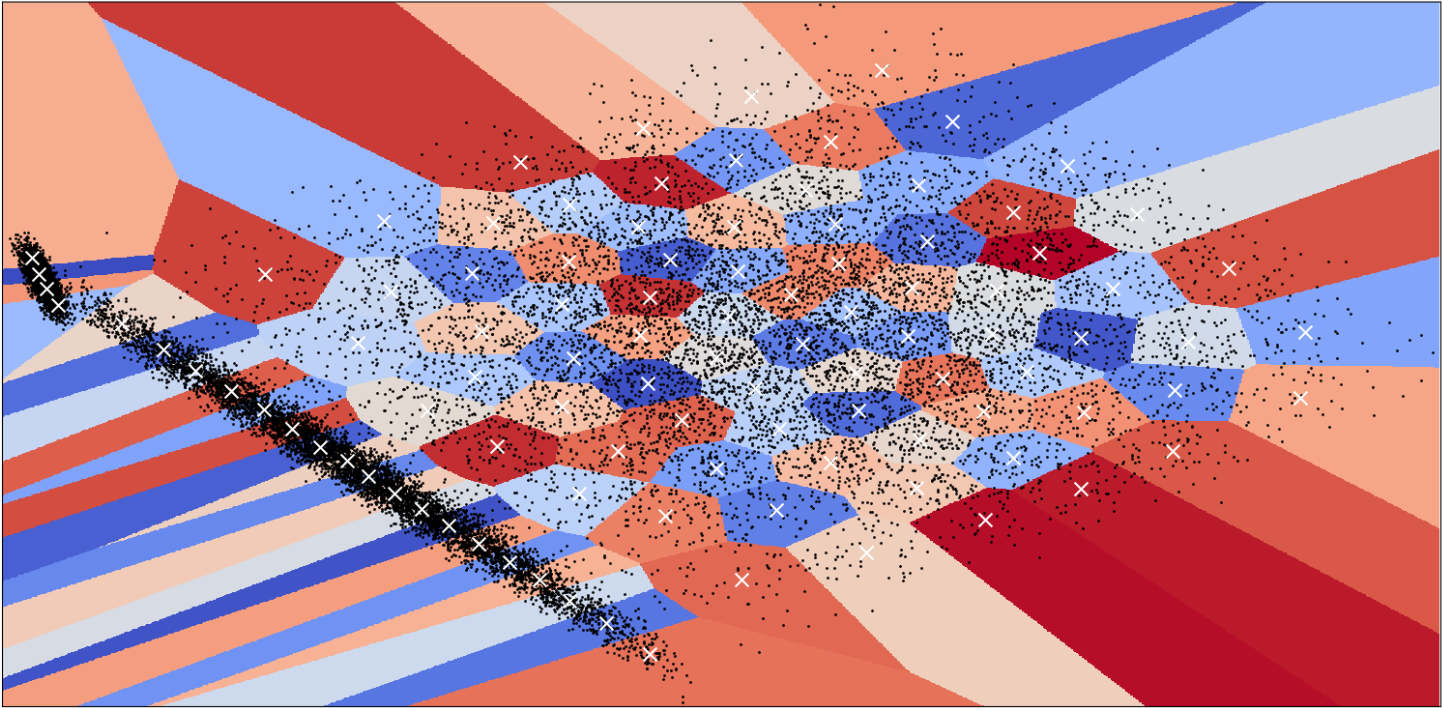}}
\caption{Example clusters of our training data using Principal Component Analysis \cite{PPCA} for visualisations in 2D, where each black dot represents a training sentence or dialogue}
\label{exampleclusters}
\end{figure}

While a small number of sentence clusters may result in actions being assigned to potentially the same cluster\footnote{In this case our system would select randomly from   sentences with the same cluster ID}, a larger number of sentence clusters would mitigate the problem, but the larger the number of clusters the larger the computational cost---i.e. more parameters in the neural net. Table~\ref{clusteredsentences} shows example outputs of our sentence clustering using 100 clusters on our training data. A manual inspection showed that while clustered sentences sometimes do not seem very similar, they made a lot of sense and they produced reasonable outputs. Our human evaluation (see Section \ref{sec:human-eval}) confirms this. All our experiments use $\mathcal{K}=100$ due to a reasonable compromise between system performance and computational expense.

The purpose of our second clustering model is to split our original training data into a group of data subsets, one subset for each ChatDQN agent in our ensemble. \textcolor{black}{We explored different numbers of clusters (20, 50, 100) and noted that the larger the number of clusters the (substantially) higher the computational expense \footnote{Our experiments ran on a cluster of 16 GPU Tesla K80, and their implementation used the following libraries: Keras (\url{https://github.com/keras-team/keras}), OpenAI (\url{https://github.com/openai}) and Keras-RL (\url{https://github.com/keras-rl/keras-rl}).}.} We chose 100 clusters for our experiments due to higher average episode rewards of cluster-based agents than non-cluster-based ones. Figure~\ref{exampleclusters} shows visualisations of our sentence and dialogue clustering using 100 clusters on our training data of 17.8K data points. A manual inspection was not as straightforward as analysing sentences due to the large variation of open-ended sets of sentences---see next section for further results.  

\subsection{Automatic Evaluation}
\label{sec:automatic-eval}
\begin{figure}[th!]
\subfigure[ChatDQN Agents -- 1 Dialogue Cluster]
{\includegraphics[width=62mm]{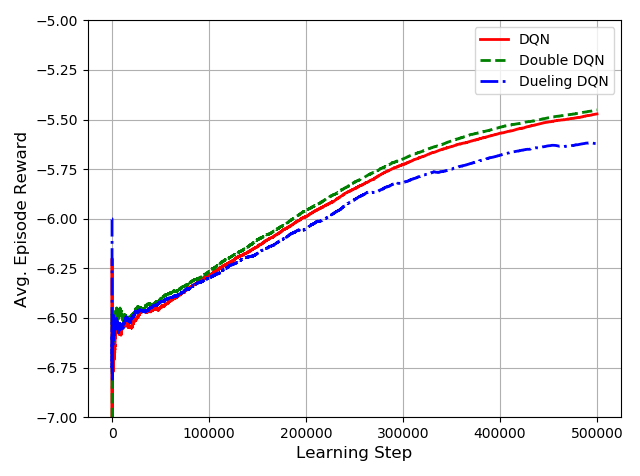}}
\subfigure[ChatDQN Agents -- 100 Dialogue Clusters]
{\includegraphics[width=62mm]{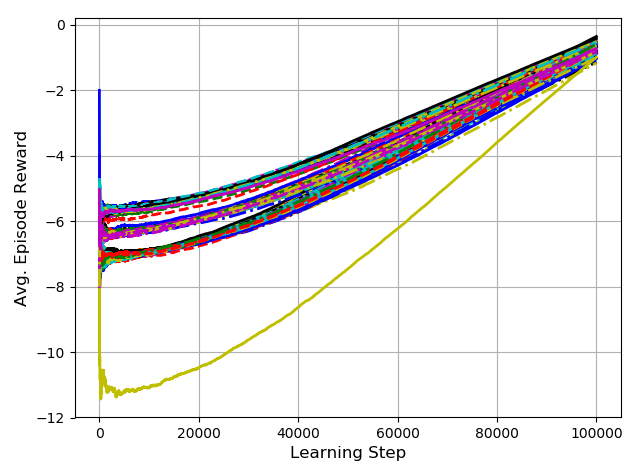}}
\vskip-5pt
\caption{Learning curves of ChatDQN agents}
\label{learningcurves}
\end{figure}

We compared three DQN-based algorithms (DQN \cite{MnihKSRVBGRFOPB15}, Double DQN \cite{HasseltGS16} and Dueling DQN \cite{WangSHHLF16}) in order to choose a baseline single agent and the learning algorithm for our ensemble of agents. The goal of each agent is to choose the human-generated sentences (actions) out of a set of candidate responses (20 available at each dialogue turn). Figure~\ref{learningcurves}(left) shows learning curves for these three learning algorithms, where we can observe that all agents indeed improve their performance (in terms of average episode reward) over time. It can also be observed that DQN and Double DQN performed similarly, and that Dueling DQN was outperformed by its counterpart algorithms. Due to its simplicity, we thus opted for using DQN as our main algorithm for the remainder of our experiments.

Figure~\ref{learningcurves}(right) shows the performance of 100 ChatDQN agents (one per dialogue cluster), where we also observe that all agents improve their performance over time. It can be noted however that the achieved average episode reward of $\sim$ -1 is much greater than that of the single agent corresponding to $\sim$ -5.5. Additional experiments reported that the lower the number of clusters the lower the average episode reward during training. We thus opted for using 100 dialogue clusters in the remainder of our experiments. 

\begin{figure}[th!]
\begin{center}
\subfigure[Avg. Episode Reward]{
\includegraphics[width=57mm]{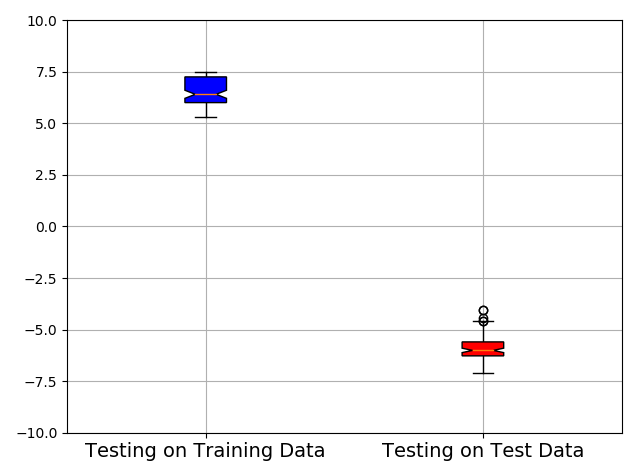}}
\hspace{-0.3cm}
\subfigure[Avg. F1 Score]{
\includegraphics[width=57mm]{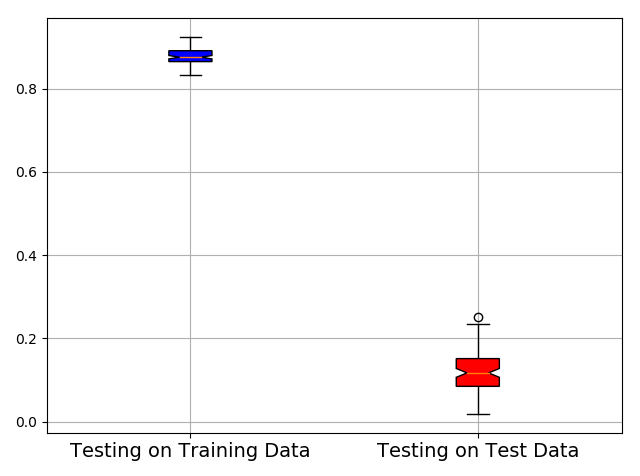}}
\hspace{-0.3cm}
\subfigure[Avg. Recall@1]{
\includegraphics[width=57mm]{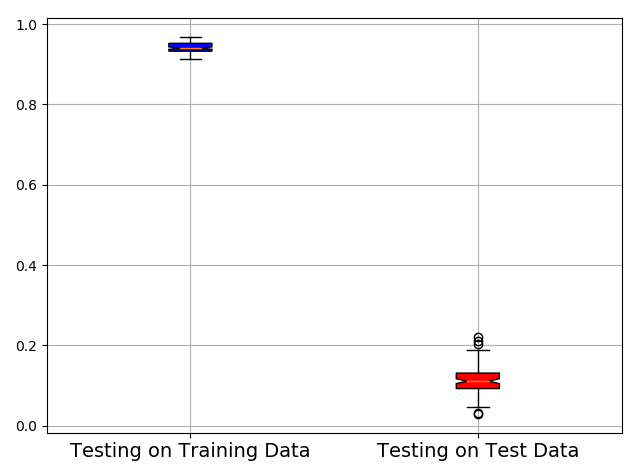}}
\hspace{-0.3cm}
\subfigure[Avg. Recall@5]{
\includegraphics[width=57mm]{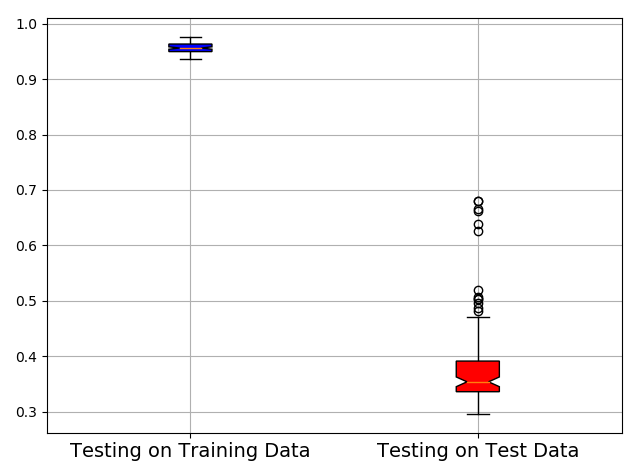}}
\end{center}
\vspace{-0.2cm}
\caption{Test performance of 100 ChatDQN agents on training (blue boxes) and test data (red boxes) using 4 evaluation metrics}
\label{boxplots}
\end{figure}

We analysed the performance of our agents further by using the test set of 999 totally unseen dialogues during training. We clustered the test set using our trained dialogue clustering model in order to assess the goodness of each agent in dialogues that were similar but not the same. The box plots in Figure~\ref{boxplots} report the performance of our DRL agents according to the following metrics while tested on training data and test data: Avg. Episode Reward, Avg. F1 score, Avg. Recall@1, and Average Recall@5. One can quickly observe the striking performance gap between testing on training data vs. testing on test data. This can be interpreted as ChatDQN agents being able to learn well how to select actions on  training data, but not being able to replicate the same behaviour on test data. This may not be surprising given that only 720 sentences (out of 263,862 training sentences and 15,586 test sentences) are shared between both sets, and it is presumably a realistic scenario seen that even humans rarely use the exact same sentences in multiple conversations. \textcolor{black}{On the one hand our results also suggest that our training dataset is rather modest, and that a larger dataset is needed for improved performance. On the other hand, our results help us to raise the question {\it `Can chitchat chatbots with reasonable performance be trained on modest datasets--- i.e. with thousands of dialogues instead of millions?'} If so, the generalisation abilities of chatbots need to be improved in future work. If not, large (or very large) datasets should receive more attention in future work on neural-based chatbots.}

Finally, we compared the performance of 5 dialogue agents on 999 dialogues with 20 candidate sentences at every dialogue turn: 
\begin{itemize}
\item {\bf Upper Bound}, which corresponds to the true human sentences in the test dataset;
\item {\bf Lower Bound}, which selects a sentence randomly from other dialogues than the one in focus;
\item {\bf Ensemble}, which selects a sentence using 100 agents trained on clustered dialogues as described in section~\ref{approach} -- the agent in focus is chosen using a regressor as predictor of dialogue reward $\hat{R}(.)$ using a similar neural net as the ChatDQN agents except for the final layer having one node and using Batch Normalisation \cite{IoffeS15} between hidden layers as in \cite{CuayahuitlEtAl2018neurips}; 
\item {\bf Single Agent}, which selects a sentence using a single ChatDQN agent trained on the whole training set; and
\item {\bf Seq2Seq}, which selects a sentence using a 2-layer LSTM recurrent neural net with attention\footnote{\url{https://github.com/facebookresearch/ParlAI/tree/master/projects/convai2/baselines/seq2seq}} -- from the Parlai framework (\url{http://www.parl.ai}) \cite{ZhangEtAl2018}, trained using the same data as the agents above.
\end{itemize}

Table~\ref{automaticeval} shows the results of our automatic evaluation, where the ensemble of ChatDQN agents performed substantially better than the single agent and Seq2Seq model.

\begin{table}[t!]
\centering
\caption{Automatic evaluation of chatbots on test data}
\begin{tabular}{|c|c|c|c|} 
\hline
Agent/Metric & Dialogue Reward & F1 Score & Recall@1 \\
\hline
\hline
Upper Bound	& 7.7800 & 1.0000 & 1.0000\\
Lower Bound	& -7.0600 & 0.0796 & 0.0461\\
Ensemble	& {\bf -2.8882} & {\bf 0.4606} & {\bf 0.3168}\\
Single Agent & -6.4800 & 0.1399 & 0.0832\\
Seq2Seq	     & -5.7000 & 0.2081 & 0.1316\\
\hline 
\end{tabular}
\label{automaticeval}
\end{table}

\subsection{Human Evaluation}
\label{sec:human-eval}
\begin{figure}[t!]
\begin{center}
\includegraphics[width=11cm]{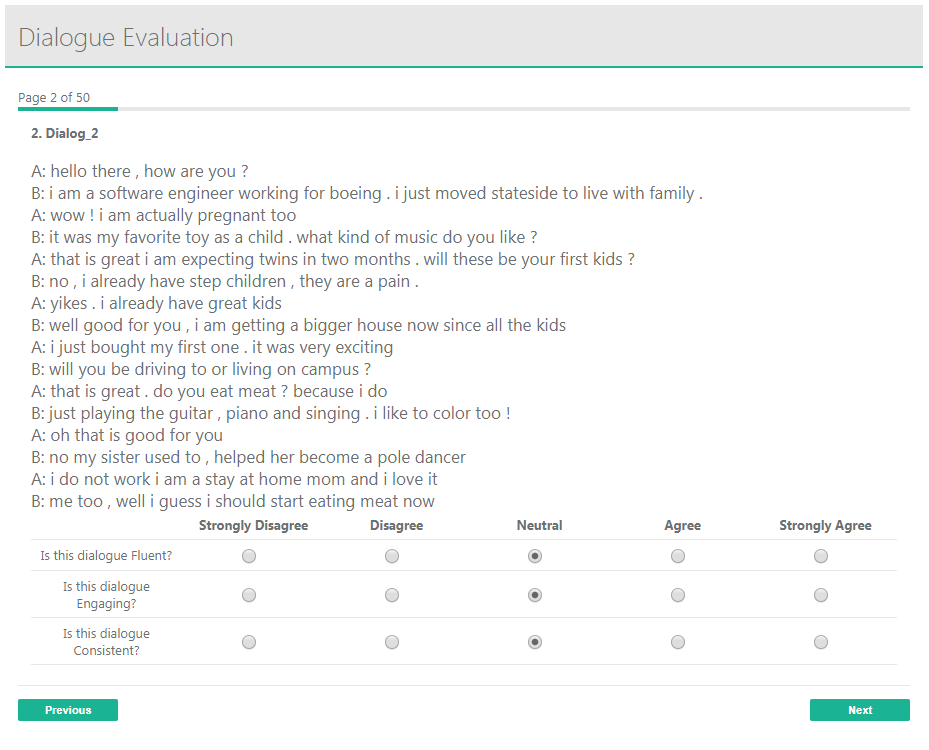}
\end{center}
\caption{Screenshot of our dialogue evaluation tool}
\label{eval_tool_page}
\end{figure}

In addition to our results above, we carried out a human evaluation using 15 human judges. Each judge was given a form of consent for participating in the study, and was asked to rate 500 dialogues (100 core dialogues---from the test dataset---with 5 different agent responses\footnote{All agents responded to the same human conversants, and they used the same sets of candidate sentences---for a fair comparison.}, dialogues presented in random order) according to the following metrics: {\it Fluency} (Is the dialogue naturally articulated as written by a human?), {\it Engagingness} (Is the dialogue interesting and pleasant to read?), and {\it Consistency} (without contradictions across sentences). This resulted in $15\times500\times3=22,500$ ratings from all judges. Figure~\ref{eval_tool_page} shows an example dialogue with ratings ranging from 1=strongly disagree to 5=strongly agree. 

\begin{figure}[th!]
\centering
\includegraphics[width=90mm]{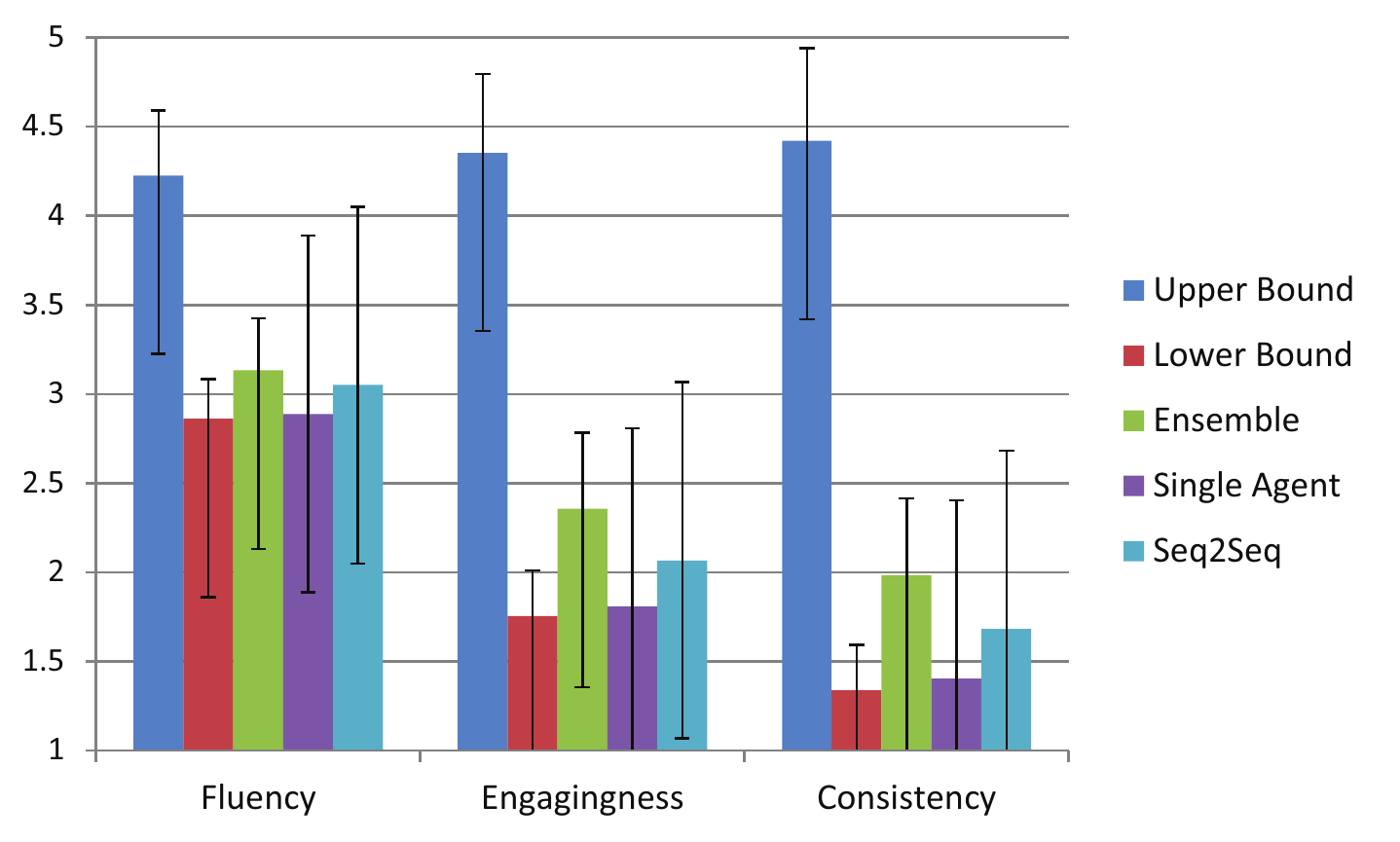}
\caption{Human evaluation results, the higher the better}
\label{barplot}
\end{figure}

Figure~\ref{barplot} shows average ratings (and corresponding error bars) per conversational agent and per metric\footnote{Note that the candidate sentences used as distractors were chosen randomly from randomly selected dialogues---which is rather challenging for action selection. Future work could consider candidate sentences from similar dialogues for potential improvements in terms of engagingness and consistency.}. As expected, the {\bf Upper Bound} agent achieved the best scores and the {\bf Lower Bound} agent the lowest scores. The ranking of our agents in Table~\ref{automaticeval} is in agreement with the human evaluation, where the {\bf Ensemble} agent outperforms the {\bf Seq2Seq} agent, and the latter outperforms {\bf Single Agent}. 
The difference in performance between the {\bf Ensemble} agent and the {\bf Seq2Seq} agent is significant at $p=0.0332$ for the Fluency metric and at $p<0.01$ for the other metrics (Engagingness and Consistency)---based on a two-tailed Wilcoxon Signed Rank Test.

\begin{figure*}[th!]
\begin{center}
\subfigure[Fluency]{
\includegraphics[width=59mm]{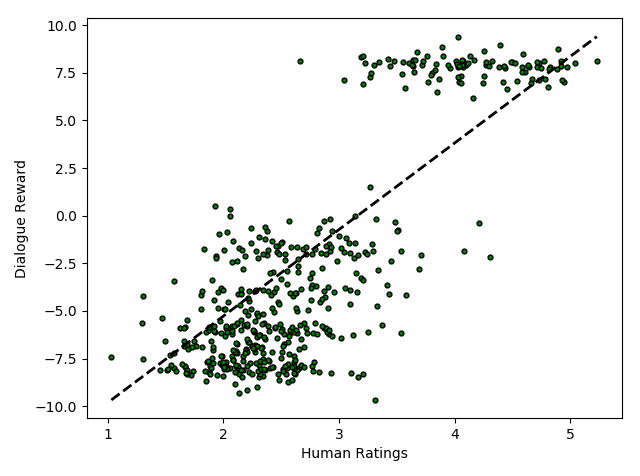}}
\subfigure[Engagingness]{
\includegraphics[width=59mm]{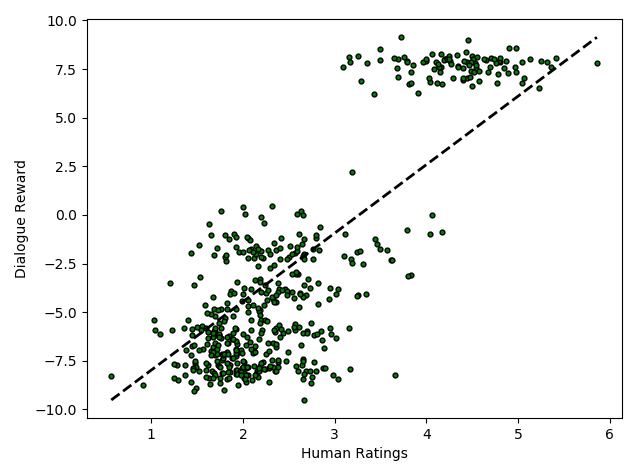}}
\subfigure[Consistency]{
\includegraphics[width=59mm]{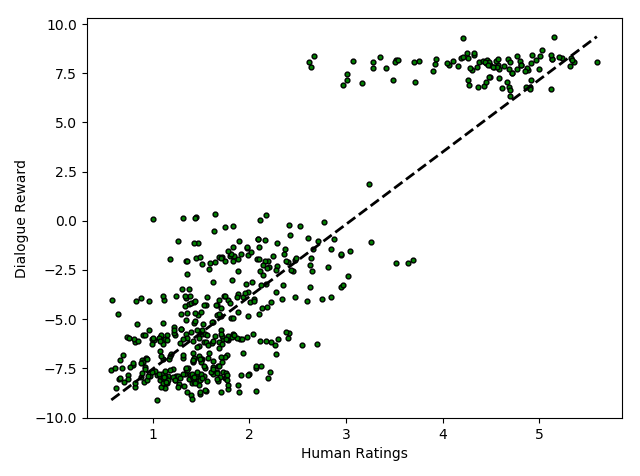}}
\vspace{-0.5cm}
\end{center}
\caption{Scatter plots showing strong correlations (with Pearson coefficients of $0.90 \le r \le 0.91$) between  predicted dialogue rewards and average human ratings as part of the human evaluation, i.e. our proposed reward function correlates with human judgements
}
\label{scatterplots2}
\end{figure*}

Furthermore, we analysed the predictive power of dialogue rewards, derived from our reward function, against human ratings on test data. This analysis revealed positive high correlations between them as shown in Figure\ref{scatterplots2}. These scatter plots show data points of test dialogues (the X-axes include Gaussian noise drawn from $\mathcal{N}(0, 0.3)$ for better visualisation), which obtained Pearson correlation scores between 0.90 and 0.91 for all metrics (Fluency, Engagingness and Consistency). This is in favour of our proposed reward function and supports its application to training open-ended dialogue agents.

\section{Conclusions and Future Work}
We present a novel approach for training Deep Reinforcement Learning (DRL) chatbots. It uses an ensemble of 100 DRL agents based on clustered dialogues, clustered actions, and rewards derived from human-human dialogues without any manual annotations. The task of the agents is to learn to choose human-like actions (sentences) out of candidate responses including human generated and randomly chosen sentences. Our ensemble trains specialised agents with particular dialogue strategies according to their dialogue clusters. At test time, the agent with the highest predicted reward is used during a dialogue. Experimental results using chitchat dialogue data report that DRL agents learn human-like dialogue policies when tested on training data, but their generalisation ability in a test set of unseen dialogues (with mostly unseen sentences, only 4.62\% seen sentences to be precise) remains a key challenge for future research in this field. As part of our study, we found the following:
\textcolor{black}{
\begin{enumerate}
\item an ensemble of DRL agents is more promising than a single DRL agent or a single Seq2Seq model---confirmed by a human evaluation; 
\item value-based DRL can be used for training chatbots---previous work mostly uses policy search methods due to infinite action sets; and 
\item our proposed reward function albeit simple was useful for training chatbots. 
\end{enumerate}
}

Future work can investigate further the proposed learning approach for improved generalisation in test dialogues. Some research avenues are as follows.
\begin{itemize}
\item Investigating other methods of sentence embedding such as \cite{DevlinCLT19,CerEtAl2018,LeMikolov2014} -- possibly with fine-tuning or domain adaptation subject to using large datasets. Other DRL algorithms such as policy search methods should also be compared or combined. In addition, other distance metrics and clustering algorithms should be used to investigate better sentence clustering and dialogue clusterings. Alternative dialogue rewards should be compared to train agents with human-like dialogue rewards across different datasets.
\item \textcolor{black}{An interesting future direction is training an ensemble of ChatDQN agents using a very large dataset -- much larger than attempted in this article. Our results seem to suggest that the larger the dataset the better generalisation on unseen data. But this requirement represents high costs in data collection and high computational expense for system training. While chatbot training using large or very large datasets is interesting, chatbot training using modest datasets is still relevant because it can save costly datasets and computational requirements.}
\item \textcolor{black}{The proposed learning approach focuses on value-based deep reinforcement learning, and it could be combined with other deep learning methods in order to investigate more effective ensembles of machine learners. For example, our ensemble of agents could include not only value-based DRL methods but also policy search methods and a variety of seq2seq methods. Although this research direction represents increased computational expense, it has the potential of showing improved performance over single agents/models. }
\item Our proposed approach in this article did not include any linguistic resources. One reason for this is the practical application of DRL agents to other languages/datasets, where linguistic resources are scarce or do not exist. Another reason is due to the fact that linguistic resources usually come at the expense of labelled data, and we aimed for investigating an approach and methodology assuming unlabelled data only. However, future work could improve the performance of DRL agents by including knowledge bases and natural language resources such as part-of-speech tagging, named entity recognition, coreference resolution, and syntactic parsing \cite{ManningEtAl2014}. 
\item \textcolor{black}{Last but not least, the proposed approach can be applied to different applications, beyond chitchat dialogue. Example applications in no particular order are as follows: combining task-oriented dialogue with open-ended dialogue \cite{Papaioannou2017hri}, strategic dialogue \cite{CuayahuitlEtAl2015nips}, spatially-aware dialogue \cite{Dethlefs17}, automatic (medical) diagnosis \cite{WeiLPTCHWD18}, in-car infotainment systems \cite{WengASHPH16}, and conversational robots \cite{CUAYAHUITL2019}, among others.}
\end{itemize}

\bibliography{refs}

\end{document}